%% file: main.tex
\pdfoutput=1
\documentclass{article}

\usepackage[dblblindworkshop,final]{neurips_2026}
\workshoptitle{Foundation Models for Temporal Systems}

\usepackage[utf8]{inputenc}
\usepackage[T1]{fontenc}
\usepackage{newunicodechar}
\newunicodechar{，}{,}
\usepackage{hyperref}
\hypersetup{hidelinks}   %
\usepackage{url}
\usepackage{booktabs}
\usepackage{multirow}
\usepackage{amsfonts}
\usepackage{amsmath}
\usepackage{amsthm}
\usepackage{microtype}
\usepackage{xcolor}
\usepackage{graphicx}
\usepackage{natbib}
\usepackage{capt-of}

\newcommand{\Bu}{\ensuremath{B_u(t)}}
\newcommand{\wiki}{\texttt{tgbl-wiki}}
\newcommand{\lastfm}{\texttt{tgbl-lastfm}}
\newcommand{\coin}{\texttt{tgbl-coin}}
\newcommand{\flight}{\texttt{tgbl-flight}}

\newcommand{\ours}{GTGT}

\title{Vocabulary-Free Next-Event Prediction on Temporal Graphs}

\author{%
  Zhishang Luo$^{1}$ \quad
  Venu Madhav Tammali$^{2}$ \quad
  Amila Weerasinghe$^{2}$ \quad
  Osso Vahabzadeh$^{2}$ \\
  \textbf{Olcay Boz}$^{2}$ \quad
  \textbf{Xue Hao}$^{2}$ \quad
  \textbf{Yusu Wang}$^{1}$ \\
  $^{1}$University of California San Diego \quad $^{2}$Amazon \\
  \texttt{\{zluo,yusuwang\}@ucsd.edu}, \texttt{\{vtammali,weera,ossovh,olcayboz,xhhao\}@amazon.com}
}

\begin{document}

\maketitle

\begin{abstract}
Autoregressive transformer training through next-token prediction has driven remarkable advances in language modeling and has become increasingly effective for time-series modeling. Yet its success in temporal graph learning remains limited, and most temporal link-prediction methods are discriminative
rather than autoregressive. While temporal graphs can have a natural serialization through chronological ordering edges, they lack an obvious graph-independent tokenization: existing node tokenization often uses a vocabulary that grows with the number of nodes, limiting scalability and cross-graph training. 
We propose \textbf{\ours{}}, \textbf{G}enerative \textbf{T}emporal \textbf{G}raph \textbf{T}ransformer,
a simple yet powerful vocabulary-free framework for self-supervised next-event prediction on temporal graphs. It predicts events for observed destinations, behavioral descriptors for novel destinations, and time gaps before new event. These targets have shared meanings across graphs, allowing one transformer to train across disjoint node sets without node-specific learned parameters. Local node identifiers let attention capture graph structure: Theoretically, we can show that our transformer-based model has the capacity for time-respecting message passing mechanisms. On Temporal Graph Benchmark (TGB), our \ours{} not only achieves comparable or better link-prediction performance than established baselines, but also has much lower training cost. 
These results open the door to temporal graph pretraining for forecasting and other downstream tasks, providing a concrete step toward temporal graph foundation models.

\end{abstract}
\section{Introduction}
\label{sec:intro}

Next-token prediction has driven advances in language models~\citep{radford2018gpt,brown2020gpt3} and recently time-series models~\citep{ansari2024chronos,liu2024timer}. Temporal graphs record events between entities over time, making their sequences a natural setting for autoregressive learning. Indeed, it is highly desirable to develop a pre-trained temporal graph foundation model, via next-token prediction, that can support link prediction, events forecasting, and other downstream tasks. 
Discriminative link-prediction training ranks destinations given a source and query time. Its representations may transfer, but the objective does not directly predict when future events occur and what future events looks like. Autoregressive next-event learning instead is more suitable for self-supervised learning.

However, one challenge for designing next-event prediction framework is to define targets
that can be shared across graphs. In language modeling, the same vocabulary can be reused across
documents~\citep{sennrich2016neural}. Therefore, a straightforward approach predicts the next destination using a vocabulary of node IDs. But in this design adding graphs will add node-specific parameters, even when nodes have similar event patterns. We therefore ask:
\emph{How to design a scalable, efficient and effective autoregressive training framework across arbitrary number of temporal graphs?}

In this paper, we propose \ours{}, a simple yet powerful vocabulary-free and self-supervised learning based framework for next-event prediction on temporal graphs. 
Specifically, our model describes a destination node by its behavioral descriptor and recent events in history if available. These kind of targets / relations are sharable across different graphs. Parameter-free local node identifiers further allow a causal Transformer to capture different graph structure without a separate graph encoder. In fact, we show that an edge-only stream (with our local node identifier) under causal attention can implement time-respecting message passing (Appendix~\ref{app:proof}).

Empirically, we following the setup of the Temporal Graph Benchmark (TGB) \citep{huang2023tgb}, and compare our model against nine baseline models (seven models from the TGB paper and two new SoTA models), including TGN~\citep{rossi2020tgn}, GraphMixer~\citep{cong2023graphmixer}, DyGFormer~\citep{yu2023dyglib}, and SimpleDyG~\citep{wu2024simpledyg}. 
Our \ours{} not only achieves comparable or better link-prediction performance, but also has much lower training cost. %
A single jointly trained model matches specialized models on three of four graphs with far fewer parameters.

Since our \ours{} is vocabulary-free with the number of parameters independent of training graphs, it can naturally scale to an increasing number of graphs. Its self-supervised
nature further allows us to train a single model that learns task-agnostic representations that can support a range of downstream tasks beyond link prediction as in our current experiments. We believe our model makes a concrete step towards building an efficient and effective temporal graph foundation model.

\par\noindent\textbf{Related work.}
SimpleDyG~\citep{wu2024simpledyg} models event sequences similar to \ours{} but using a learned item vocabulary that grows with node sets sizes. DyGFormer~\citep{yu2023dyglib} avoids a learned node vocabulary, but scores candidate pairs through discriminative training instead of generative next-token training. MiNT~\citep{shamsi2024mint} trains temporal GNNs across graph snapshots to predict graph-level properties. \ours{} instead learns autoregressively from individual events. For structural encoding, TokenGT~\citep{kim2022tokengt} uses orthonormal identifiers with node and edge tokens 
on static graphs. \ours{} uses only edge tokens with causal temporal attention and is specifically designed for temporal graphs.
Our node identifiers also relate to random node features, known to
increase the expressive power of message-passing~\citep{sato2021random,abboud2021surprising}.

\section{Method}
\label{sec:method}

A temporal graph is a sequence of events $e_i=(u_i,v_i,t_i,a_i)$,
where $u_i$ and $v_i$ are the source and destination, $t_i$ is the
timestamp, and $a_i$ is the edge type. 
Given a source $u$ and the observed graph history, our goal is to predict the next event as a new token. 
At link-prediction evaluation, the source, query
time, and candidate destinations are supplied, so our autoregressive next-event prediction can directly adapt to link-prediction without further training. 
Below we describe our proposed model \textbf{G}enerative \textbf{T}emporal \textbf{G}raph \textbf{T}ransformer ({\bf \ours{}}); see Figure \ref{fig:arch}. Details see Appendix~\ref{app:repro}.

\begin{figure}[t]
\centering
\includegraphics[width=0.88\linewidth]{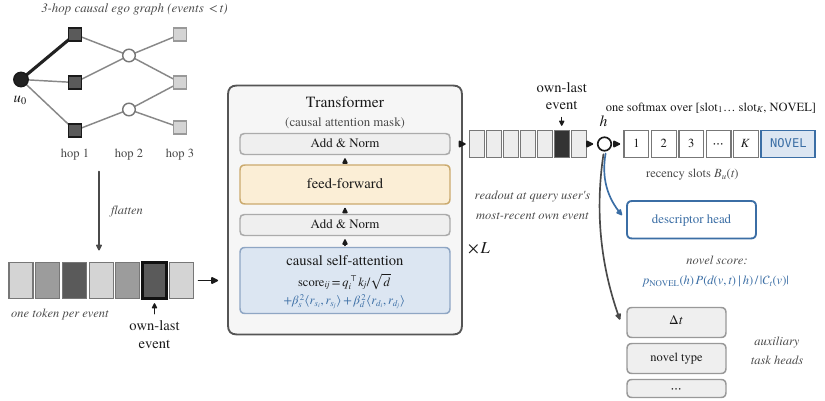}
\caption{\ours{} predicts the next destination and time gap from event histories.
Local identifiers let causal attention match shared endpoints without a learned node vocabulary.}
\label{fig:arch}
\end{figure}

\par\noindent\textbf{Serialization, encoder, and readout.}
\label{sec:encoder}
\label{par:readout}
For each query source, we sample a three-hop ego-graph with at most
512 events up to the source's latest event, so the context never depends on the query time, and order them by timestamp: each resulting event sequence is the {\it context} of \ours{}. Causal attention follows this
sequence order, including equal-timestamp events. Each event becomes one token containing its edge type, inter-event
time gap, destination's recency position, and behavioral descriptor. Here, {\it recency} refers to how recently a destination interacted
with the source node. We order distinct past destinations from most to
least recently observed in the source's history, and a destination's rank in this list is its
{\it recency position}.
The descriptor summarizes the destination's past activity, including
popularity and recency. These fields use shared categories rather than
learned node embeddings. Appendix~\ref{app:token} defines the fields,
and Appendix~\ref{app:repro} gives the sampling procedure.
For source $u$, the contextual representation $h$ of its latest event
token predicts its next events. During teacher forcing, each
supervised token predicts its source's true next event, which need not be in the context. All predictions are computed
in parallel in one forward pass, giving multiple training targets
per sampled window.

\par\noindent\textbf{Vocabulary-free structural identifiers.}
\label{sec:identity}
Behavioral descriptors alone are not sufficient to identify shared nodes and graph topology. For each node
$v$ in a context (an event sequence), we sample a {\it random node identifier}, which is a random unit vector $z_v = g_v/\|g_v\|_2$ with $g_v\sim\mathcal{N}(0,I_{d})$, that is, a vector drawn uniformly from the unit sphere in $\mathbb{R}^{d}$; $d = 32$ in our experiments. 
Each occurrence of $v$ within the same context uses the same identifier. For new context we resample identifiers and thus nothing needs to be recorded across sequences. 
These identifiers are injected to each layer of the transformer. The key utility is that these vectors are near orthonormal. Their inner-products help us match shared source / destination nodes within transformers, thereby allowing attention to aggregate events with shared endpoints and capturing graph topology. In Appendices~\ref{app:proof} and~\ref{app:synth} we show theoretically these identifiers allow us to implement standard time-respecting message passing mechanism as in TGAT~\citep{xu2020tgat}; and provide empirical evidence that with these identifiers, our \ours{} captures local graph structures.  

\par\noindent\textbf{Next-event targets and training.}
\label{sec:factorization}
\label{sec:objectives}
Given a source $u$, query time $t$, and its next-event prediction
representation $h$, we score each candidate destination $v$. Let $\Bu$ be the capped list of distinct destinations previously
observed for $u$, ordered from most to least recently observed. Destinations in $\Bu$ are recurrent; all others
are novel, including previously seen nodes outside the list.
We build a {\it pointer head} as a single softmax predicts over $K$ recency slots and \textsc{Novel},
with probabilities $p_j(h)$ and $p_{\mathrm{nov}}(h)$.
We also predict destination descriptors for novel events
and time-gap buckets for all events. A small zero-initialized head adds a
per-slot logit bias from each slot's history statistics (repeat count,
time since last event, mean interval), conditioned on $h$, and a second one
adds a bias to the time-gap logits from the source's own past gap-bucket
histogram.
Training uses all supervised positions in each window;
See details settings in Appendix~\ref{app:token}.

\par\noindent\textbf{Link prediction from next-event outputs.}
The pointer, as we explained above, and descriptor outputs give the candidate score for link prediction that used at test time:
\begin{equation}
\label{eq:main_score}
s(v\mid u,t,h)=
\begin{cases}
\log p_{j(v)}(h), & v\in\Bu,\\[2pt]
\log p_{\mathrm{nov}}(h)+\log D(v,t\mid h)
-\log|\mathcal{C}_t(v)|, & v\notin\Bu,
\end{cases}
\end{equation}
where $j(v)$ is the position of $v$ in $\Bu$,
$D(v,t\mid h)$ is the product of its predicted descriptor-block
probabilities, and $\mathcal{C}_t(v)$ contains all destinations
sharing its descriptor at time $t$. Since several destinations may have same behavioral descriptor, during link-prediction we divide the predicted probability of that descriptor by the number of destinations sharing it, $|\mathcal{C}_t(v)|$. Time-gap predictions do not enter this score.
We rank TGB~\citep{huang2023tgb} candidates directly,
without additional link-prediction training.

\section{Results}
\label{sec:results}
\label{sec:transfer}

\par\noindent\textbf{Evaluation protocol.}
We follow the same set up as TGB~\citep{huang2023tgb}, using their chronological splits and official evaluation candidates. We exclude \texttt{tgbl-review} and \texttt{tgbl-comment} because of internal legal
restrictions associated with the industry affiliation of some authors of this paper. We therefore evaluate on the remaining three TGB link-prediction datasets: namely, \wiki{}, \flight{}, and \coin{}, plus an additional publicly available dataset \lastfm{}.

We compare against nine baseline methods, where seven of them are directly evaluated in TGB \cite{huang2023tgb} paper, and two new transformer based methods DyGFormer~\cite{yu2023dyglib} and SimpleDyG~\cite{wu2024simpledyg}. 
Trainable parameter vary across methods and datasets, see Table~\ref{tab:params}.

Table~\ref{tab:transfer} includes the strongest baselines and representative
controls across heuristics, message passing, and Transformer-based methods.
Table~\ref{tab:transfer_full} in Appendix~\ref{app:baselines} gives the full comparison.
Table~\ref{tab:transfer} shows that \ours{} achieves the comparable or better Mean Reciprocal Rank (MRR, higher values better) among all compared methods in 3 out 4 datasets, while on \lastfm{}, it exceeds all other methods other than DyGFormer. Because these graphs are highly recurrent, a no-parameter recency control is
strong ($0.817$ on \wiki{}, $0.833$ on \flight{}, $0.783$ on \coin{}); \ours{}
exceeds it on all four graphs, by a small margin on \wiki{} and \coin{} and by $0.17$ on
\lastfm{}. 
We also train a single joint model instead of one model per corpus. 
The joint model, with $1.13$M parameters against $3.93$M across four
specialists, matches the specialists on \wiki{}, \flight{}, and \coin{} but
falls behind on \lastfm{} ($0.213$ vs.\ $0.333$), where each graph receives a
quarter of a $10$k-step budget.  
We further test transfer to unseen graphs: a model trained on three graphs,
without any further training, ranks links on the held-out fourth graph.
Without ever seeing the test graph, it already outperforms DyGFormer and
GraphMixer trained on that graph on \wiki{}, \flight{}, and \coin{}; on \flight{}
it is comparable to the model trained on \flight{} itself ($0.888$ vs.\ $0.884$)
(Appendix~\ref{app:logo}).

\input{editorial/tables/transfer}

Importantly, our model requires much less training time. Measured in the
$6$k-step configuration, and compared
with the previous best-performing model DyGFormer, training four
\ours{} specialists takes $86.3$ minutes and training one joint model
takes $47.7$ minutes, compared with $564.8$ minutes for four DyGFormer models.
Figure~\ref{fig:efficiency} (Appendix~\ref{app:cost}) plots each model by its
avg. test MRR and total training time: our two models achieve the highest
average MRR with the least training time among the models shown.
For inference, averaged over the four datasets, inference takes
$0.94$\,ms per link for \ours{} specialists, compared with
$4.84$\,ms for TGN, $45.64$\,ms for GraphMixer, and $20.32$\,ms
for DyGFormer. See Appendix~\ref{app:cost} for the timing setup and per-dataset results.

\par\noindent\textbf{Prediction beyond links.}
The model also predicts the next time gap over 13 buckets
(Appendix~\ref{app:event_prediction}). We show that the model's predictions
outperform standard parameter-free baselines on \wiki{}, \lastfm{}, and \coin{}, including
a per-user Poisson process, but only slightly outperform the per-user empirical
gap distribution, we therefore do not consider time-gap prediction a main result but instead as a possible future direction for a foundation model. 
Component ablations appear in Appendix~\ref{app:analysis}.

\paragraph{Conclusion.}
In this paper we develop \ours{}, a simple yet powerful next-event prediction framework for temporal graphs, and demonstrate its effectiveness and efficiency over link-prediction tasks. Our vocabulary-free formulation allows additional graphs to contribute training data without adding node-specific learned parameters, and our zero-shot results on unseen graphs already outperform several baselines trained on them. This makes \ours{} a good candidate for building a temporal graph foundation model: for future work, we will expand training to larger, more diverse graph mixtures and evaluate multi-step forecasting and downstream tasks.

\begin{ack}
This work was conducted in collaboration with Amazon, and all GPU computation was
supported by funding from Amazon.
\end{ack}

\bibliographystyle{plainnat}
\bibliography{references}
\clearpage
\appendix

\section{Representation and training objective}
\label{app:axes}

\ours{} combines autoregressive self-supervised next-event prediction
with vocabulary-free tokenization. SimpleDyG~\citep{wu2024simpledyg}
uses a learned item vocabulary for autoregressive prediction, while
DyGFormer~\citep{yu2023dyglib} avoids a learned node vocabulary but
uses discriminative link prediction. Our contribution is a
tokenization and target formulation that makes vocabulary-free
next-event prediction possible. Recency positions and behavioral
descriptors provide shared targets across graphs without requiring
node-specific learned parameters. These two axes, the use of a learned
node vocabulary and the choice of a generative versus discriminative
objective, separate SimpleDyG and DyGFormer from our model.

Our motivation is to build a training framework that can learn from
increasingly large and diverse temporal graph datasets.
Link prediction is one evaluation of this framework; the model
is trained to predict next-event components, including timing.
We aim to use this shared training task as a basis for future
foundation models that support forecasting, more downstream
tasks, and generalize to new graph datasets. 

Our main argument is that a temporal graph foundation model should be
trained as a next-event generative model, not only through link
prediction. Once the target is the
next event, tokenization determines
both the input representation and the prediction target, and it must map
different temporal graphs into a shared space so that one model can train
across datasets with disjoint node sets. Event tokenization is therefore
central to cross-graph pretraining, yet it remains largely under-explored.
Our formulation is one design point along the two axes above; we view
improvements to tokenization, attention, and training objectives as the
main directions for making large-scale temporal graph pretraining
effective and practical.

\section{Controlled same-node prediction}
\label{app:synth}
\label{sec:analysis}
\label{sec:probes}

A model trained with next-event prediction encodes some graph structure: linear
probes on its frozen states recover whether two events share a source or a
destination (Appendix~\ref{app:aggregation}, Table~\ref{tab:probes}). However,
an untrained network with the same architecture, and even the input token
embeddings, already reach high AUC on these probes, so frozen probes cannot
attribute the encoded structure to any one component. To
isolate the effect of the node identifier, we therefore use a
controlled task, trained and evaluated on the same objective: a binary
classification that predicts whether a pair of events share a node. Here we
remove all meaningful node content, so that the identifier is the only signal
that can tie events of the same node together (details below).
Table~\ref{tab:identity_necessity} reports this task on the four datasets,
adding the time channel and the identifiers one at a time.

Without either time or identifiers, AUC is near chance ($0.494$--$0.506$).
Identifiers alone raise it to $0.962$--$0.986$. Time alone gives a smaller
improvement, largest on \wiki{} ($0.494$ to $0.656$). These results show
that identifiers provide an effective signal for same-node matching in
this controlled task. Appendix~\ref{app:aggregation} separately examines
frozen-representation probes.

\input{editorial/tables/identity_necessity}

The four settings are the four dataset graphs, using their real event streams
and topology. In place of the removed node content, each event receives an
independent Gaussian feature vector, so its content does not encode a persistent
node identity. When enabled, identifiers are sampled anew for each
context. Negative event pairs are matched to positive pairs by positional gap.
The time channel uses normalized event timestamps. These controls aim to reduce
alternative ways to predict the label.

The near-chance results with neither channel support the intended control. Time-only AUC above chance on \wiki{} also shows
that temporal information remains useful in at least one setting.

\section{Context selection and frozen-representation probes}
\label{app:aggregation}

The structural task uses same-node labels for training. Frozen probes instead
ask which information can be recovered after next-event training. These
experiments answer different questions.

\paragraph{Information from the attention window.}
Full attention over $N$ events costs $O(N^2)$, making it impractical for the
$67$M-event \flight{} stream. Sampling a temporal ego-neighborhood limits this
cost, as neighborhood sampling does for inductive
GNNs~\citep{hamilton2017graphsage}. It can also provide structural information:
nearby queries may include overlapping events. This context selection is
distinct from explicitly matching endpoints with identifiers. We do not assign
a formal $1$-WL bound to the full sampling-and-attention pipeline without a
separate analysis in this paper.

\paragraph{Frozen probes.}
Unlike Appendix~\ref{app:synth}, where the model is trained end-to-end on a
same-node task, here linear probes read node information from the frozen states
of models trained only for next-event prediction, with the model's real token
fields as input. As discussed in Appendix~\ref{app:synth}, under this setting there is no guarantee that next-event prediction task require model to encode graph structure,  so the results are more likely test under this pretraining objective and with our design of node identifier, can the model still properly encode graph structure.
Table~\ref{tab:probes} reports the probes for the per-corpus models of
Table~\ref{tab:transfer}. For each dataset, we encode $400$ test contexts and
split them into two halves, one to fit the probes and one to evaluate them. For
two tokens $i$ and $j$ of the same context, a logistic-regression probe reads
$[h_i\odot h_j,\,|h_i-h_j|]$ and predicts whether the two events share a source
or a destination (co-reference). Neighbour retrieval ranks the other tokens of the
context for a given token with a probe of the same form; a token is relevant if it
comes from a different source that shares a destination with the given token's
source in the context. We apply the same probes to an untrained network with the
same architecture and to the input token embeddings.

The trained states encode source identity beyond both controls on all four
datasets (AUC $0.957$, $0.832$, $0.705$, and $0.936$, against at most $0.914$,
$0.685$, $0.623$, and $0.826$). Destination identity is already present in each
token's input fields, namely the recency position, the descriptor, and the
identifier: the controls reach $0.87$--$0.95$, and training lowers destination
AUC on \lastfm{}, \flight{}, and \coin{} while raising it slightly on \wiki{}.
Neighbour retrieval is low and close to the controls except on \wiki{} ($0.212$
vs.\ $0.162$), so the frozen states recover endpoint identity more readily than
the two-hop neighbourhood. These probes assess local node information rather
than multi-hop composition.
\input{editorial/tables/probes}

\input{proof.tex}

\section{Token construction and prediction targets}
\label{app:token}

\paragraph{Recency slots.}
For source $u$, $\Bu$ contains at most $K$ distinct destinations observed before
time $t$, ordered by their most recent events with $u$. Slot $1$ denotes
the most recent destination, slot $2$ the second-most recent distinct destination,
and so on. A destination outside this list is novel under the model's
definition, even if it appeared earlier. This definition differs from whether
a source--destination pair has ever occurred.

\paragraph{Behavioral descriptors.}
The descriptor $d(v,t)$ summarizes a destination's history before $t$.
Its fields are log-bucketed degree, per-type events counts, the mean and
coefficient of variation of inter-event gaps, age since the first event, and
time since the most recent event. The fields are quantized into one-hot buckets
and concatenated. Two destinations with the same quantized statistics receive
the same descriptor. Each descriptor block has its own softmax prediction head.
Appendix~\ref{app:repro} specifies the bin edges and descriptor width.

\paragraph{Training objective.}
The reported models use a unified softmax over recency slots and a
\textsc{Novel} slot. The recurrent probability is the total mass on real
slots, rather than the output of a separate sigmoid head. The five active
loss terms are
\begin{align}
\mathcal{L}={}&\mathcal{L}_{\mathrm{act}}+\mathcal{L}_{\mathrm{ptr}}
+0.5\mathcal{L}_{\mathrm{gap}}\nonumber\\
&+0.5\mathcal{L}_{\mathrm{type}}+0.5\mathcal{L}_{\mathrm{desc}}.
\label{eq:encoded_objective}
\end{align}
The action term $\mathcal{L}_{\mathrm{act}}$ is binary cross-entropy for the
recurrent-versus-novel decision. The pointer term $\mathcal{L}_{\mathrm{ptr}}$
uses cross-entropy on recurrent positions. The gap term
$\mathcal{L}_{\mathrm{gap}}$ uses cross-entropy over discretized time-gap
buckets at all positions. The event-type term $\mathcal{L}_{\mathrm{type}}$ and
descriptor term $\mathcal{L}_{\mathrm{desc}}$ apply only to novel positions;
event type uses categorical cross-entropy. Descriptor loss averages the target
negative log-probabilities over blocks and novel positions:
\begin{equation}
\mathcal{L}_{\mathrm{desc}}^{(\mathrm{pos})}
=-\frac{1}{|\mathcal{I}_{\mathrm{nov}}|B}
\sum_{i\in\mathcal{I}_{\mathrm{nov}}}\sum_{b=1}^{B}
\log P_{\theta,b}(d_{i,b}\mid h_i),
\end{equation}
where $\mathcal{I}_{\mathrm{nov}}$ is the set of supervised novel positions
and $B$ is the number of descriptor blocks. Each loss term in
Equation~\ref{eq:encoded_objective} is applied at two kinds of positions. The
first is the query position, the source's latest event token, whose prediction
is the one used at evaluation. The second is every other supervised token in the
window: during teacher forcing we feed the ground-truth history and, in the same
forward pass, also ask each such token to predict its own next event, and we add
these per-token losses with weight $1.0$. In both cases the target of a token is
the next event that shares its source, which need not be the immediately
following token in the sequence.

\paragraph{Candidate ranking.}
The ranking score uses the unified pointer and descriptor likelihood:
\begin{equation}
\label{eq:factorization}
s(v\mid u,t,h)=
\begin{cases}
\log p_{j(v)}(h),&v\in\Bu,\\[3pt]
\displaystyle\log p_{\mathrm{nov}}(h)
+\sum_{b=1}^{B}\log P_{\theta,b}(d_b(v,t)\mid h)
-\log|\mathcal{C}_t(v)|,&v\notin\Bu.
\end{cases}
\end{equation}
Class sizes are computed exactly over all items in the relevant corpus,
using their descriptors as of query time $t$. They are not estimated from
the supplied candidate set. Candidates with identical descriptors share the
class mass equally. The gap and novel-event-type predictions do not enter
the ranking score as separate factors. We use these values as ranking
scores without assuming that their exponentials sum to one over named nodes.
Generating a descriptor also does not by itself generate a complete graph
event: it still needs to be assigned to an existing or new node.

\section{Experimental settings}
\label{app:repro}

\paragraph{Model settings.}
The reported per-dataset models use four causal Transformer-encoder layers,
$d_{\mathrm{model}}=128$, four attention heads, dropout $0.1$, and feed-forward
width $512$. Each layer uses pre-LayerNorm. The unified pointer has $256$
recency slots on \wiki{}, \lastfm{}, and \flight{}, and $1{,}024$ on \coin{},
plus the \textsc{Novel} slot. The model also has gap, novel-type, and
descriptor-block heads.
With $256$ slots, a model has $931{,}094$ learned parameters, and the
pointer-bias head adds $2{,}961$ ($934{,}055$, the reported \wiki{},
\lastfm{}, and \flight{} models); the gap-history head adds $299$.
With $1{,}024$ slots (\coin{} and the joint model), it has $1{,}131{,}431$.
The reported per-corpus models (Table~\ref{tab:transfer}) train for $6$k steps
on \wiki{}, $20$k on \lastfm{}, and $10$k on \flight{} and \coin{}, with seeds
$0$, $1$, and $2$ (we report mean $\pm$ standard deviation); the joint model trains
for $10$k steps with seed $0$. \flight{} and \coin{} are evaluated
on $200$k-query subsets of the test split, the others on the full test split. Parameter count depends on slot
capacity and configured features, but not on an entity-indexed learned table.

\paragraph{Identifier implementation and theoretical scope.}
The identifier dimension is $d=32$ ($d_P$ in Appendix~\ref{app:proof}). Each identifier is a standard
Gaussian vector normalized to unit length, as assumed by the near-orthogonality
argument in Remark~\ref{rem:coherence}. Identifiers are drawn per example,
carry no learned parameters, and enter only through attention-score biases.
More than 32 node vectors in this space cannot all be mutually orthogonal.
The exact orthonormal construction in Appendix~\ref{app:proof} is therefore
an idealized capacity result, with a separate extension for near-orthogonal
identifiers; it does not establish exact orthogonality of the implemented codes.

\paragraph{Optimization and checkpoint selection.}
The reported checkpoints use AdamW with $\beta_1=0.9$, $\beta_2=0.999$,
learning rate $3\times10^{-4}$, weight decay $0.01$, and batch size $64$.
Training runs for a fixed number of steps ($6{,}000$ on \wiki{}, $20{,}000$ on
\lastfm{}, and $10{,}000$ on \flight{} and \coin{} for the reported models;
$6{,}000$ for the base configuration) with a OneCycle learning-rate schedule
peaking at $3\times10^{-4}$. We validate every $2{,}000$ steps and at the end on
$2{,}000$ validation queries, scoring each validation by the sum of pointer MRR on
recurrent queries and recurrent/novel accuracy, and restore the best-scoring
checkpoint for evaluation. The joint model uses the same schedule, with each
batch split evenly across its training graphs and one slot capacity
($1{,}024$) for all graphs.

\paragraph{Gap and descriptor bins.}
Bin counts equal the number of boundaries plus one. Time quantities are
measured in seconds. The gap boundaries are used for inter-event gap tokens,
descriptor mean gaps, age, and recency. The boundaries are:
\begin{center}
\small
\begin{tabular}{lp{0.62\linewidth}r}
\toprule
Field & Bin boundaries & Bins \\
\midrule
Gap / age / recency &
$0,\allowbreak 1,\allowbreak 10,\allowbreak 60,\allowbreak 300,\allowbreak 1800,\allowbreak 3600,\allowbreak 21600,\allowbreak 86400,\allowbreak 604800,\allowbreak 2592000,\allowbreak 31536000$ & 13 \\
Degree &
$0,\allowbreak 1,\allowbreak 2,\allowbreak 4,\allowbreak 8,\allowbreak 16,\allowbreak 32,\allowbreak 64,\allowbreak 128,\allowbreak 256,\allowbreak 512,\allowbreak 1024,\allowbreak 4096,\allowbreak 16384$ & 15 \\
Gap CV & $0,0.25,0.5,0.75,1.0,1.5,2.0,3.0$ & 9 \\
Type count & $0,1,2,4,8,16,64,256$ & 9 \\
\bottomrule
\end{tabular}
\end{center}
The descriptor concatenates one-hot blocks for degree, mean gap, gap CV,
age, recency, and a count block for each edge type. For $T$ edge types,
its width is $15+13+9+2\cdot13+9T=63+9T$, or $72$ for these single-type
datasets. These definitions correspond to the reported implementation in
\texttt{v3/descriptors.py}.

\paragraph{Context sampling and supervision.}
We first retrieve the root node's 128 most recent events. We select its
20 most recently interacting distinct items and retrieve each item's
16 most recent events. We then expand the second-hop nodes with four
most recent events per node, subject to a total context cap of 512 events.
This allocates more history to the root and less to more distant nodes.
Sampled events form one timestamp-ordered sequence. Equal-timestamp events
are randomly permuted each time the context is used during training.
Causal attention follows this sequence order. Teacher-forced targets are
constructed by pairing each supervised token with the next event with the
same source, rather than the immediately following event across all sources;
this is the source's true next event in the training stream, which need not lie
in the context. All quotas are filled with events up to the source's latest
event, never up to the query time (Appendix~\ref{app:event_prediction}).
At evaluation, the query source's latest observed event token provides
the prediction readout. The same sampler is used in training and in every
evaluation.
Training uses eight cross-batch negatives in the pointer softmax; these do
not change the evaluation score in Equation~\ref{eq:factorization}.

\paragraph{Data and candidates.}
We use \texttt{py-tgb} $2.3.0$, official chronological TGB splits,
and the official presampled evaluation negatives. Each query ranks one positive
against $999$ negatives on \wiki{} and \lastfm{} ($1{,}000$ candidates in total),
and against $20$ negatives on \flight{} and \coin{} ($21$ in total).

\paragraph{Evaluation history.}
All historical information used to score a query is strictly earlier than
its target timestamp. Within the input context, events with equal timestamps
follow arrival order: a token may attend to earlier-arriving tokens at the
same timestamp, but not to later-arriving ones.

\paragraph{Metric.}
For a positive score $s^+$ and negative scores $(s^-_k)_{k=1}^{m}$, define
$n_>=\sum_k\mathbf{1}[s^-_k>s^+]$ and
$n_\geq=\sum_k\mathbf{1}[s^-_k\geq s^+]$.
The average-tie rank is $r=1+(n_>+n_\geq)/2$, and MRR averages $1/r$ over
queries. A descriptor collision therefore produces a tie, not a zero score.
Assigning zero reciprocal rank to out-of-list destinations would instead
define a coverage-adjusted metric, which is distinct from candidate-ranking MRR.
We use the TGB evaluator for this average-tie candidate-ranking MRR.

\paragraph{Ablation and baseline settings.}
Each ablation in Table~\ref{tab:ablation} starts from the per-corpus model,
trained for $6$k steps, and removes one component; all other settings,
including the sampler and the evaluation candidates, are unchanged. Without the
recency pointer, every training target is treated as a novel destination, and
we keep the final checkpoint, since the selection score depends on the pointer.
The SimpleDyG baseline is our implementation following the original
paper~\citep{wu2024simpledyg} for our datasets. It uses node identities on both
input and output, with the same causal backbone and a $40$K-step budget, and it is
not an official leaderboard entry.

\paragraph{Joint training.}
The joint model trains for $10$K steps with batch size $64$; each batch contains
$16$ examples from each of the four datasets, so every dataset contributes $160$K
examples. Recency lists, descriptors, and evaluation candidates stay local to each
graph. The shared slot capacity is the largest dataset-specific cap ($1{,}024$),
with unused slots masked for each example. The joint model uses the same
architecture and heads as the per-corpus models. Each per-corpus model sees $384$K
examples on \wiki{}, $1.28$M on \lastfm{}, and $640$K on \flight{} and \coin{},
so per-dataset exposure is not matched, especially on \lastfm{}. The leave-one-out
models use the same schedule, with each batch split evenly across their three
training datasets.

\section{Full baseline comparison}
\label{app:baselines}

Table~\ref{tab:transfer_full} reports the full baseline roster, combining
local runs and reported leaderboard values. Our per-corpus results are averaged
over three seeds, with a standard deviation of at most $0.002$.
\input{editorial/tables/transfer_full}

\paragraph{Recurrent and novel queries.}
Because these graphs are highly recurrent, we also split the test queries by
whether the true destination is in the source's recency list
(Table~\ref{tab:breakdown}). \ours{} improves over the recency control in
different ways across datasets. On \lastfm{}, it ranks recurrent destinations much
better than recency ($0.381$ vs.\ $0.189$), so most of its gain comes from
ordering past destinations rather than from their recency alone. On \flight{}, the
gain also comes from recurrent destinations ($0.931$ vs.\ $0.872$), while recency
is better on novel queries ($0.278$ vs.\ $0.183$): a destination outside the
source's recency list may still have been seen with the source long ago, and
recency still scores it by the time of that event. On \coin{}, both are equal on
recurrent queries ($0.913$), and \ours{} ranks novel destinations better
($0.317$ vs.\ $0.283$). On \wiki{}, both subsets stay close to recency,
consistent with the small overall margin in Table~\ref{tab:transfer}.
\input{editorial/tables/breakdown}

\section{Transfer to unseen graphs}
\label{app:logo}
Joint training on four graphs shares one model across them, but every test graph
was also seen in training. To test transfer to a graph the model has never seen,
we train one model on three of the four graphs and evaluate it zero-shot on the
fourth, for each choice of held-out graph (Table~\ref{tab:logo}). No parameter
is specific to a graph, so the held-out graph needs no new parameters: its
events are tokenized with the same shared fields and scored with
Equation~\ref{eq:main_score}.
Table~\ref{tab:logo} shows that our model transfers to graphs it has never
seen. On \flight{}, the zero-shot model reaches $0.888$, comparable to the model
trained on \flight{} itself ($0.884$) and substantially higher than the recency
control ($0.833$) and DyGFormer ($0.843$). On \wiki{} and \coin{}, the zero-shot model
reaches $0.805$ and $0.766$, which is within $0.02$--$0.03$ of the model trained
on the target graph and higher than DyGFormer ($0.780$, $0.747$) and
GraphMixer ($0.519$, $0.760$) trained on it. On these two graphs, however, it
remains slightly below the no-parameter recency control ($0.805$ vs.\ $0.817$ on
\wiki{}, $0.766$ vs.\ $0.783$ on \coin{}). On \lastfm{}, transfer is more difficult: the
zero-shot model reaches $0.151$, compared with $0.333$ when training on \lastfm{}
itself, and is also below recency ($0.166$). We expect that training on more and
more diverse graphs, rather than three, will reduce these gaps, and we leave it
for future work.

\input{editorial/tables/logo}

\section{Dataset structure and history coverage}
\label{app:corpora}

Pair recurrence ranges from $0.831$ to $0.966$
(Table~\ref{tab:graphstats}). Repeated events are therefore a large part
of the evaluated prediction task, making recency heuristics important controls.
This is different from coverage by a capped recent-history list.
Table~\ref{tab:answerable} measures that coverage on the test split at cap $256$.
An out-of-list destination can have prior activity elsewhere, or can be an old
destination of the source that has fallen out of the cap. It need not be a
globally new node.

\wiki{} and \lastfm{} are bipartite and have $1$K destinations each.
\flight{} and \coin{} use a shared source/destination node space. The flight
stream records $67$M events at about $1.4$K distinct timestamps, making the
handling of simultaneous events especially important. \coin{} has $638{,}486$
destinations, a setting where a learned entity table can dominate model size.

The standard evaluation candidate widths are $1{,}000$ on \wiki{} and
\lastfm{}, and $21$ on \flight{} and \coin{}. MRR should be compared between
methods within a dataset; differences across datasets also reflect the number
and construction of evaluation candidates.

\input{editorial/tables/graphstats}
\input{editorial/tables/corpora}
\input{editorial/tables/answerable}

\section{Parameter count, training time, and runtime}
\label{app:cost}
\begin{figure}[t]
\centering
\begin{minipage}[t]{0.6\linewidth}
\vspace{0pt}
\includegraphics[width=\linewidth]{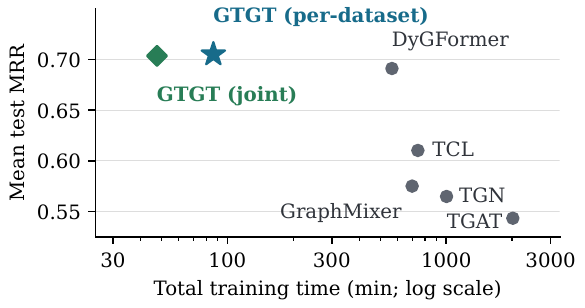}
\caption{Mean test MRR versus total training time across four datasets, for
the $6$k-step configuration.}
\label{fig:efficiency}
\end{minipage}
\end{figure}

\paragraph{Learned parameters.}
Four independently trained specialists contain approximately $3.93$M learned
parameters in total, compared with $1.13$M for the joint model. Thus, the joint
model uses under one-third of the aggregate learned parameters required by
four specialists. This comparison concerns model parameters, not training
compute or graph-specific storage.
At a fixed configuration, our learned parameter count is independent of the
number of entities. The model still stores graph-specific histories and
behavioral statistics, so total storage need not be constant. The learned-vocabulary
baseline grows from $940{,}776$ parameters on the $1$K-destination datasets to
$83{,}176{,}470$ on \coin{}, according to the main table's provenance.
Table~\ref{tab:params} reports the reference configuration and neural baselines.

\paragraph{Training time.}
Table~\ref{tab:runtime} reports the training-time.
All baselines ran on a device with same GPU L40S, and the TGB baseline methods are run with official TGB protocol, codes and implementation, where our model is trained using our own implementation. For next-event prediction, each training batch contains $64$ sampled contexts of up to $512$ events, and every supervised token in a context provides a training target in the same forward pass. Contexts are sampled on CPU threads in parallel with GPU training, and attention uses the standard scaled dot-product attention in PyTorch, with the identifier inner products added as an attention bias. 

Our training-time estimates are $13.9$, $13.5$, $34.6$, and $24.3$ minutes
on \wiki{}, \lastfm{}, \flight{}, and \coin{}, respectively.
The baselines use early stopping on validation, while our models train for a fixed number of steps and keep the best validation checkpoint (Appendix~\ref{app:repro}); different methods has different data input format and training samplers implementations, so runtime comparisons should be
interpreted together with the training schedule and resulting accuracy.

\paragraph{Inference.}
Table~\ref{tab:cost} measures positives ranked per second, including the full
candidate set, on an idle L40S. Our model scores $930$ positives/s on \wiki{}
and $896$ on \lastfm{}, above the listed baselines. On \flight{}, TGN and
DyRep exceed our $2{,}295$ positives/s. On \coin{}, TCL exceeds our $898$
positives/s. Thus, the measured advantage depends on dataset, candidate count,
and slot capacity. These measurements do not establish a uniform inference
speedup or isolate candidate count as the cause.

\input{editorial/tables/params}
\input{editorial/tables/runtime}
\input{editorial/tables/cost}

\section{Component ablations and candidate-width sweep}
\label{app:analysis}

\paragraph{Component ablations.}
Table~\ref{tab:ablation} removes one component at a time from the per-corpus
model, with the same tokenization, sampler, and $6$k-step budget. The recency
pointer is the most important component: without it, MRR drops from $0.823$ to
$0.472$ on \wiki{} and from $0.251$ to $0.092$ on \lastfm{}. Removing the
slot-bias head lowers MRR on both datasets ($0.819$ and $0.229$). Removing
local identifiers or descriptor prediction changes MRR only slightly
($0.247$ and $0.244$ on \lastfm{}, no change on \wiki{}), since on these
highly recurrent datasets most test destinations are ranked by the pointer;
the role of local identifiers is examined separately in
Appendix~\ref{app:synth}.

\paragraph{Candidate-count analysis.}
Table~\ref{tab:negsweep} reports the candidate-width sweep. The largest
setting uses $999$ negatives and one positive, giving $1{,}000$ candidates
in total; smaller widths keep a random subset of the official negatives.

\input{editorial/tables/ablation}
\input{editorial/tables/negsweep}

\section{Next-event component prediction}
\label{app:event_prediction}

We evaluate the per-corpus models of Table~\ref{tab:transfer} (seed $0$) on
$20{,}000$ test queries from each dataset, sampled among queries whose source has
at least one earlier event. Because queries from new sources are excluded, the
share of recurrent queries is higher than in Table~\ref{tab:breakdown} on \wiki{}
($92.7\%$ vs.\ $89.2\%$).
The recency-slot capacity is $256$ for \wiki{}, \lastfm{}, and
\flight{}, and $1{,}024$ for \coin{}.

\par\noindent\textbf{Metrics.}
The recurrent/novel decision and time-gap prediction are evaluated
on all sampled test positions. A destination is recurrent when it
appears in the retained recency list. Pointer accuracy and MRR are
computed only on recurrent positions and rank recency slots;
they differ from link-prediction MRR over TGB's supplied candidates.
Event-type accuracy is computed only on novel positions.
Time prediction uses the $13$ predefined gap buckets in
Appendix~\ref{app:repro}, rather than continuous-time regression.

These metrics evaluate next-event queries, not every teacher-forced token
in the input window. During training, each supervised token predicts the
next event sharing its source, which need not be the adjacent token.

\par\noindent\textbf{Time-gap baselines.}
Table~\ref{tab:event_gap_full} compares the time-gap predictions with standard
parameter-free baselines. Each baseline outputs a distribution over gap buckets,
using only the source's own events up to $t_{\text{last}}$, the same history the
model may use; gaps are taken between distinct timestamps, since the target gap is
always positive. Random is uniform over the $13$ buckets. Majority is the bucket
distribution over training queries. Persistence predicts the source's previous gap
bucket. The Poisson process models the source's gaps as exponential with the
source's mean gap and gives each bucket its probability under this distribution. The per-user
empirical distribution is the frequency of each bucket among the source's past
gaps. Persistence, the Poisson process, and the empirical distribution are mixed
with or smoothed toward Majority so that every bucket has non-zero probability;
we select each weight on $10{,}000$ validation queries and evaluate once on the test queries. We
report negative log-likelihood (NLL) and bucket accuracy on the same
$10{,}000$ test queries per dataset.

\input{editorial/tables/event_gap_full}

\par\noindent\textbf{What the model sees.}
The gap head reads the same representation $h$ used for link prediction. Its
inputs are the context tokens up to the source's latest event $t_{\text{last}}$,
whose only timing information is their own past inter-event gap buckets; the
gap-history head adds the source's gap-bucket histogram up to $t_{\text{last}}$.
Neither the query time $t$ nor the elapsed time $t-t_{\text{last}}$ enters any
input. A
regression test checks that the readout's inputs and gap logits are identical for
every query time after $t_{\text{last}}$.

\par\noindent\textbf{Why the context ends at the source's latest event.}
The time-gap target is $t-t_{\text{last}}$, so no input to the readout may depend
on the query time $t$. If the context were collected at $t$, each node's quota
would be filled with its most recent events before $t$. Events between
$t_{\text{last}}$ and $t$ are hidden from the readout by causal attention, but they
would still take up part of the quotas, leaving fewer earlier events for the
readout to see. The number and kind of events before the readout would then reveal
the gap. We therefore collect every context up to $t_{\text{last}}$. For the same reason, each teacher-forced target is its source's
true next event in the training stream rather than the next event of that source
inside the context, which would skip events of partially sampled sources. Link
prediction may still use the query time: it enters the candidate scores through
the slot statistics and descriptors computed at $t$.

\begin{table}[t]
\centering
\small
\setlength{\tabcolsep}{4pt}
\caption{\textbf{Complete next-event component metrics.}
All values are test accuracies except pointer MRR.
Pointer metrics are conditional on recurrent events.
Time-gap results are in Table~\ref{tab:event_gap_full}.}
\label{tab:event_full}
\begin{tabular}{lrrrr}
\toprule
& \multicolumn{2}{c}{Recurrent/novel}
& \multicolumn{2}{c}{Pointer}\\
\cmidrule(lr){2-3}
\cmidrule(lr){4-5}
Dataset & Majority & Model & Accuracy & MRR  \\
\midrule
\wiki{}   & 0.927 & 0.940 & 0.883 & 0.922 \\
\lastfm{} & 0.879 & 0.889 & 0.299 & 0.380 \\
\flight{} & 0.938 & 0.938 & 0.209 & 0.309 \\
\coin{}   & 0.795 & 0.846 & 0.449 & 0.561 \\
\bottomrule
\end{tabular}
\end{table}

\par\noindent\textbf{Interpretation.}
The recurrence decision improves over the majority baseline on
\wiki{}, \lastfm{}, and \coin{}, while matching it approximately
on \flight{} (Table~\ref{tab:event_full}). High recurrence accuracy
alone therefore does not establish improvement over a constant predictor.
For time-gap prediction (Table~\ref{tab:event_gap_full}), the model learns
useful timing information: its NLL is well below Random, Majority,
Persistence, and the Poisson process on \wiki{}, \lastfm{}, and \coin{}. Without
the gap-history head, it outperforms the per-user empirical distribution on
\wiki{} but not on \lastfm{} and \coin{}. Adding the gap-history head, which
gives the readout the source's gap-bucket histogram directly, makes the model
outperform all baselines on all three datasets, by $0.05$, $0.06$, and $0.03$
nats over the per-user empirical distribution, without hurting link prediction
(\lastfm{} MRR $0.262$ vs.\ $0.251$ at $6$k steps; \coin{} $0.794$ vs.\
$0.791$ at $10$k steps). We therefore regard this as a modest
improvement over parameter-free baselines rather than a strong time-prediction
result. Most remaining errors occur at session boundaries: on \lastfm{}, $36\%$
of next events occur more than $30$ minutes later, and the model predicts the
correct bucket for only $7\%$ of these. Even an oracle that
predicts each user's most frequent test-period gap reaches only $45\%$ accuracy,
so much of the gap is unpredictable from history. On \flight{}, $99.7\%$ of targets fall in one bucket, so we do not compare
with the baselines there. Event-type accuracy is trivial because each
evaluated dataset contains only one edge type.

These measurements evaluate individual next-event components,
rather than generated event sequences. The gap and novel-event-type
outputs are trained but do not enter the link-ranking score as
separate factors. Descriptor losses in the training logs are
optimization diagnostics; held-out descriptor likelihood and
complete event-stream generation are not evaluated here.

\end{document}

%% file: editorial/tables/transfer.tex
\begin{table}[t]
\centering
\small
\setlength{\tabcolsep}{4pt}
\caption{\textbf{Link prediction on TGB} (test Mean Reciprocal Rank, or MRR, higher is better). 
Results for all nine baselines are in Appendix~\ref{app:baselines}.
Per-corpus uses separate models; joint shares one model. SimpleDyG is our
implementation following the original paper. Bold marks the highest MRR.
\ours{} per-corpus: mean $\pm$ std over three seeds; joint: seed $0$; budgets and test sets in Appendix~\ref{app:repro}.}
\label{tab:transfer}
{%
\begin{tabular}{lrrrr}
\toprule
Method & \texttt{tgbl-wiki} & \texttt{tgbl-lastfm} & \texttt{tgbl-flight} & \texttt{tgbl-coin} \\
\midrule
  EdgeBank$_{tw}$   & 0.571 & 0.141 & 0.387 & 0.580 \\
  Recency (no parameters)    & 0.817 & 0.166 & 0.833 & 0.783 \\
  Popularity (no parameters) & 0.040 & 0.013 & 0.603 & 0.607 \\
  TGN        & 0.611 & 0.220 & 0.791 & 0.637 \\
  GraphMixer & 0.519 & 0.249 & 0.772 & 0.760 \\
  \midrule
  DyGFormer  & 0.780 &  \textbf{0.395} & 0.843 & 0.747 \\
  SimpleDyG & 0.583 & 0.184 & 0.785 & 0.534 \\
\midrule
  \ours{} (per-corpus)  & \textbf{0.823}\,{\scriptsize$\pm$0.000} & 0.333\,{\scriptsize$\pm$0.002} & 0.884\,{\scriptsize$\pm$0.001} & \textbf{0.793}\,{\scriptsize$\pm$0.002} \\
  \ours{} (joint, single model)   & 0.822 & 0.213 & \textbf{0.892} & \textbf{0.793} \\
\bottomrule
\end{tabular}
}
\end{table}

%% file: editorial/tables/identity_necessity.tex
\begin{table}[htbp]
\centering
\small
\setlength{\tabcolsep}{4pt}
\caption{\textbf{Same-node prediction with time and local identifiers.}
AUC on the content-free task, with each channel enabled or disabled.
The \emph{Neither} column, with no time and no identifier, is the control:
the label is a pure structure function, so with both time and identifiers removed, AUC is near chance. ($0.5$).
Mean $\pm$ standard deviation over three seeds.}
\label{tab:identity_necessity}{%
\begin{tabular}{lrrrr}
\toprule
Dataset & Neither & $+$Time only & $+$Identity only & $+$Time $+$Identity \\
\midrule
\wiki{}            & 0.494 $\pm$ 0.007 & 0.656 $\pm$ 0.014 & 0.982 $\pm$ 0.001 & \textbf{0.991 $\pm$ 0.001} \\
\lastfm{}          & 0.506 $\pm$ 0.009 & 0.524 $\pm$ 0.009 & 0.962 $\pm$ 0.010 & \textbf{0.968 $\pm$ 0.017} \\
\texttt{tgbl-coin} & 0.501 $\pm$ 0.019 & 0.542 $\pm$ 0.003 & \textbf{0.976 $\pm$ 0.004} & 0.971 $\pm$ 0.005 \\
\flight{}          & 0.506 $\pm$ 0.006 & 0.510 $\pm$ 0.010 & \textbf{0.986 $\pm$ 0.001} & 0.963 $\pm$ 0.007 \\
\bottomrule
\end{tabular}
}
\end{table}

%% file: editorial/tables/probes.tex
\begin{table}[t]
\centering
\small
\setlength{\tabcolsep}{5pt}
\caption{\textbf{Linear probes on frozen representations} of the seed-$0$
per-corpus models of Table~\ref{tab:transfer}. Co-reference: AUC on $4{,}000$
balanced token pairs from held-out contexts; neighbour retrieval: recall@$10$
over $1{,}000$ queries. \emph{Untrained} applies the same probes to a network with
the same architecture and random weights; \emph{input embeddings} applies them to
the token embeddings before the first layer.}
\label{tab:probes}
\begin{tabular}{llrrrr}
\toprule
Probe & States & \wiki{} & \lastfm{} & \flight{} & \coin{} \\
\midrule
\multirow{3}{*}{Source co-reference}
  & trained          & \textbf{0.957} & \textbf{0.832} & \textbf{0.705} & \textbf{0.936} \\
  & untrained        & 0.914 & 0.664 & 0.623 & 0.787 \\
  & input embeddings & 0.876 & 0.685 & 0.561 & 0.826 \\
\midrule
\multirow{3}{*}{Destination co-reference}
  & trained          & \textbf{0.945} & 0.752 & 0.886 & 0.912 \\
  & untrained        & 0.931 & \textbf{0.886} & \textbf{0.914} & \textbf{0.946} \\
  & input embeddings & 0.888 & 0.867 & 0.907 & 0.945 \\
\midrule
\multirow{3}{*}{Neighbour retrieval}
  & trained          & \textbf{0.212} & \textbf{0.029} & \textbf{0.027} & 0.055 \\
  & untrained        & 0.162 & 0.025 & 0.025 & \textbf{0.065} \\
  & input embeddings & 0.163 & 0.023 & 0.023 & 0.061 \\
\bottomrule
\end{tabular}
\end{table}

%% file: proof.tex
\providecommand{\R}{\mathbb{R}}
\providecommand{\ind}{\mathbf{1}}
\providecommand{\inn}[1]{\langle #1\rangle}
\providecommand{\In}{\mathcal{I}}
\providecommand{\Out}{\mathcal{O}}
\providecommand{\stream}{\mathcal{S}}
\providecommand{\avg}{\operatorname{avg}}
\ifcsname c@theorem\endcsname\else
  \newtheorem{theorem}{Theorem}
  \newtheorem{lemma}[theorem]{Lemma}
  \newtheorem{corollary}[theorem]{Corollary}
  \theoremstyle{definition}\newtheorem{definition}[theorem]{Definition}
  \theoremstyle{remark}\newtheorem{remark}[theorem]{Remark}
  \theoremstyle{plain}
\fi

\section{Time-respecting message passing with identifier-based attention}
\label{app:proof}

This appendix gives the construction and proof behind the expressivity claim of
Section~\ref{sec:identity}. The result establishes sufficient representational
capacity: one attention-and-MLP block can implement one round of the message
passing defined below. Content weights are shared across graphs. For finite
softmax attention, the routing strength needed for a chosen error can depend
on the stream length and feature bounds.

\subsection{Setting}

\par\noindent\textbf{Temporal edge stream.}
A temporal graph is an event stream $\stream=(e_1,\dots,e_n)$,
$e_i=(u_i,v_i,x_i)$, with source $u_i\in V$, target $v_i\in V$, and an edge
attribute $x_i\in\R^{d_e}$ (edge type, an encoding of the inter-event gap,
and any other per-event field). Timestamps are non-decreasing in $i$; ties are
fixed by the serialisation, and everything below is indexed by position $i$.
Nodes carry features $X\colon V\to\R^{d_X}$ (possibly $X\equiv0$). Nothing is
assumed about $V$: nodes may appear as both sources and targets.

\par\noindent\textbf{Causal neighbourhoods.}
For a node $w$ and position $i$,
\[
\In_i(w)=\{\,j\le i : v_j=w\,\},\qquad
\Out_i(w)=\{\,j\le i : u_j=w\,\}.
\]
For a finite index set $S$ and vectors $(y_j)$ write
$\avg_{j\in S}y_j=\frac{1}{1+|S|}\sum_{j\in S}y_j$, the mean over $S$
together with one zero vector; $\avg_\emptyset=0$.

\begin{definition}[Time-respecting message passing]\label{def:mpnn}
Fix $L\ge1$ and ReLU MLPs $\psi_k,\psi'_k\colon\R^{d_X+d_e}\to\R^{d_m}$ and
$\phi_k\colon\R^{d_X+2d_m}\to\R^{d_X}$, $k=1,\dots,L$. Let $h^{(0)}_w(i)=X_w$ and, for
$k\ge1$ and every node $w$ and position $i$,
\begin{align}
a^{(k)}_w(i)&=\avg_{j\in\In_i(w)}\ \psi_k\!\big(h^{(k-1)}_{u_j}(j),\,x_j\big),
\label{eq:a}\\
b^{(k)}_w(i)&=\avg_{j\in\Out_i(w)}\ \psi'_k\!\big(h^{(k-1)}_{v_j}(j),\,x_j\big),
\label{eq:b}\\
h^{(k)}_w(i)&=\phi_k\!\big(h^{(k-1)}_w(i),\,a^{(k)}_w(i),\,b^{(k)}_w(i)\big).
\label{eq:h}
\end{align}
\end{definition}

$h^{(k)}_w(i)$ is the state of $w$ \emph{as of position $i$}: it depends only
on events at positions $\le i$, and the message along event $j$ carries the
sender's state as of $j$, the position of the interaction. Equal timestamps
still follow the chosen serialization order. Assigning every event the same
timestamp therefore does not reduce this recursion to ordinary synchronous
static message passing.

\par\noindent\textbf{Provenance and scope of Definition~\ref{def:mpnn}.}
Definition~\ref{def:mpnn} uses neighbor states at their interaction positions,
with a zero-augmented mean aggregator. Related temporal models use
interaction-time histories, including TGAT and causal
walks~\citep{xu2020tgat,wang2021cawn}. The relevant distinction is between
monotone temporal histories and access to every neighbor's state at a later
query time~\citep{souza2022provably}. Here, dependency positions are
non-increasing when traced away from the query; equal positions are allowed
because the neighborhoods use $j\le i$. The theorem establishes containment
only for the stated recursion. It does not establish containment of every
TGAT or TGN configuration.

\par\noindent\textbf{Orthonormal identifiers.}
Each node $w$ has an identifier $r_w\in\R^{d_P}$, $\|r_w\|=1$, with
$\inn{r_a,r_b}=0$ for distinct nodes $a\ne b$ occurring in $\stream$.
(This requires $d_P\ge$ the number of distinct nodes in the window; the
near-orthogonal case is handled in Remark~\ref{rem:coherence}.) Identifiers are
redrawn per sequence; the construction never refers to a particular draw.

\par\noindent\textbf{Edge-only causal transformer.}
There is one token per event and no node tokens. Token $i$ carries a content
vector $z_i\in\R^{D}$ and the pair of endpoint identifiers $(r_{u_i},r_{v_i})$, which
are \emph{not} part of $z_i$. A stream is prefixed with a BOS token
$i=0$ with $z_0=0$. A layer is $H$ causally masked attention heads followed by a
position-wise ReLU MLP, each with a residual connection (normalisation layers
omitted). Head $\eta$ has content projections $W^\eta_Q,W^\eta_K,W^\eta_V$, a
scalar $\beta_\eta\ge0$, and an endpoint pairing
$(p_\eta,q_\eta)\in\{u,v\}^2$. For $1\le j\le i$,
\begin{equation}\label{eq:qk}
\mathbf q^\eta_i=\big[\,W^\eta_Q z_i\;\big\|\;\beta_\eta\, r_{p_\eta(i)}\big],\qquad
\mathbf k^\eta_j=\big[\,W^\eta_K z_j\;\big\|\;\beta_\eta\, r_{q_\eta(j)}\big],
\end{equation}
where $p_\eta(i)\in\{u_i,v_i\}$ according to $p_\eta$, likewise $q_\eta(j)$.
For this construction, the zero-value BOS entry is assigned logit
$\beta_\eta^2$ for every query. It therefore matches the score of a true
identifier match when the content query and key projections are zero.
This is a specified null-attention entry, not the key of an ordinary fixed
BOS token. The score for event tokens is
$\inn{\mathbf q^\eta_i,\mathbf k^\eta_j}$ for $j\le i$ and $-\infty$ for $j>i$.
No learned matrix acts on an identifier, and identifiers never enter values.
The layer MLP may have enough depth to compose the message and update
functions below. Thus, a layer here means an attention-and-MLP block; the
claim does not impose the usual fixed two-linear-map feed-forward architecture.

\begin{remark}[Pairings and the implementation]
Concatenating both identifiers onto both query and key with one scalar per block
gives the sum of the two pairings $(u,u)$ and $(v,v)$: ``same source'' and
``same target''. The construction below also uses $(u,v)$ and $(v,u)$,
obtained by swapping the key's two identifier blocks in that head. This is a
fixed permutation, not a parameter, and preserves every statement below. On a
bipartite stream (sources and targets disjoint) the cross pairings are
never needed.
\end{remark}

\subsection{One layer, one round}

\begin{lemma}[Identifier routing]\label{lem:routing}
Let head $\eta$ have $W^\eta_Q=W^\eta_K=0$ and $\beta^2_\eta=\alpha$, and let
$M_i=\{\,1\le j\le i : q_\eta(j)=p_\eta(i)\,\}$. Then the attention weights at
query $i$ are equal on $M_i\cup\{0\}$, and the total weight on
$\{1,\dots,i\}\setminus M_i$ is at most $i\,e^{-\alpha}$. Consequently, if
$\|\mathbf v_j\|\le B$ for all values and $\mathbf v_0=0$,
\[
\Big\|\sum_{j\le i}w_{ij}\mathbf v_j-\avg_{j\in M_i}\mathbf v_j\Big\|\le 2B\,i\,e^{-\alpha}.
\]
\end{lemma}

\begin{proof}
Scores equal $\alpha\inn{r_{p_\eta(i)},r_{q_\eta(j)}}$, which is $\alpha$ for
$j\in M_i\cup\{0\}$ and $0$ otherwise by orthonormality. Each off-set weight is
therefore at most $e^{-\alpha}$ times any on-set weight, and on-set weights are
at most $1$; summing over at most $i$ off-set positions gives the mass bound
$\delta\le i e^{-\alpha}$. Since $\mathbf v_0=0$, the on-set part of the output
is exactly $(1-\delta)\avg_{M_i}\mathbf v_j$ and the off-set part is $\delta$
times a convex combination of values, so the deviation is at most
$\delta\,\|\bar{\mathbf v}-\avg_{M_i}\mathbf v_j\|\le2B\delta$.
\end{proof}

The match set is the causal neighbourhood of the query's selected endpoint:
\begin{align*}
(v,v)&\colon M_i=\In_i(v_i), &
(u,u)&\colon M_i=\Out_i(u_i),\\
(u,v)&\colon M_i=\In_i(u_i), &
(v,u)&\colon M_i=\Out_i(v_i).
\end{align*}
The BOS token is what makes $\avg$ (rather than the plain mean) appear and
keeps the softmax well defined when $M_i=\emptyset$.

\par\noindent\textbf{Slots.} Partition the content as
\[
z_i=\big[\,s_i\;\|\;d_i\;\|\;x_i\;\|\;\omega^{\mathrm{in}}_i\;\|\;\omega^{\mathrm{out}}_i\;\|\;c_i\,\big],
\qquad s_i,d_i\in\R^{d_X},\ \ \omega^{\mathrm{in}}_i,\omega^{\mathrm{out}}_i\in\R^{d_m},\ \ c_i\in\R^{4d_m},
\]
read as: state of the source, state of the target, edge attribute, the message
this event contributes as an in-event of its target, the message it
contributes as an out-event of its source, and scratch space for received
aggregates. A node's state thus has no token of its own; it is held in the
$s$-slot of its out-events and the $d$-slot of its in-events.

\begin{definition}[Round invariant]
Layer $k$ \emph{holds round $k$} if for every $i\ge1$
\[
s_i=h^{(k)}_{u_i}(i),\qquad d_i=h^{(k)}_{v_i}(i),\qquad
\omega^{\mathrm{in}}_i=\psi_{k+1}\big(h^{(k)}_{u_i}(i),x_i\big),\qquad
\omega^{\mathrm{out}}_i=\psi'_{k+1}\big(h^{(k)}_{v_i}(i),x_i\big),
\]
with the convention $\psi_{L+1}=\psi'_{L+1}=0$.
\end{definition}

\begin{lemma}[One layer advances one round]\label{lem:step}
Suppose the input to layer $k$ holds round $k-1$. Then there is a choice of the
four heads and the MLP of layer $k$, independent of the identifiers and of $\stream$,
such that in the hard-attention limit $\alpha\to\infty$ the output of layer $k$
holds round $k$.
\end{lemma}

\begin{proof}
Use four heads with $W_Q=W_K=0$, $\beta^2=\alpha$, and value projections that
select one slot:
\[
\begin{array}{lll}
\text{head T:}& \text{pairing }(v,v),& W_V\text{ selects }\omega^{\mathrm{in}},\\
\text{head S:}& \text{pairing }(u,u),& W_V\text{ selects }\omega^{\mathrm{out}},\\
\text{head T$'$:}& \text{pairing }(u,v),& W_V\text{ selects }\omega^{\mathrm{in}},\\
\text{head S$'$:}& \text{pairing }(v,u),& W_V\text{ selects }\omega^{\mathrm{out}}.
\end{array}
\]
By Lemma~\ref{lem:routing} and the round-$(k-1)$ invariant, their outputs at
token $i$ are
\begin{align*}
\text{T:}\quad&\avg_{j\in\In_i(v_i)}\psi_k\big(h^{(k-1)}_{u_j}(j),x_j\big)=a^{(k)}_{v_i}(i),
&\text{S:}\quad&\avg_{j\in\Out_i(u_i)}\psi'_k\big(h^{(k-1)}_{v_j}(j),x_j\big)=b^{(k)}_{u_i}(i),\\
\text{T$'$:}\quad&\avg_{j\in\In_i(u_i)}\psi_k\big(h^{(k-1)}_{u_j}(j),x_j\big)=a^{(k)}_{u_i}(i),
&\text{S$'$:}\quad&\avg_{j\in\Out_i(v_i)}\psi'_k\big(h^{(k-1)}_{v_j}(j),x_j\big)=b^{(k)}_{v_i}(i),
\end{align*}
by \eqref{eq:a}--\eqref{eq:b}. The output projection writes these four vectors
into the scratch slot $c_i$ through the residual connection. The token now
contains every argument of the two updates
\[
h^{(k)}_{u_i}(i)=\phi_k\big(s_i,\,a^{(k)}_{u_i}(i),\,b^{(k)}_{u_i}(i)\big),\qquad
h^{(k)}_{v_i}(i)=\phi_k\big(d_i,\,a^{(k)}_{v_i}(i),\,b^{(k)}_{v_i}(i)\big)
\]
of \eqref{eq:h}. The layer MLP computes both, then
$\psi_{k+1}(h^{(k)}_{u_i}(i),x_i)$ and $\psi'_{k+1}(h^{(k)}_{v_i}(i),x_i)$, and
writes them to $s,d,\omega^{\mathrm{in}},\omega^{\mathrm{out}}$ while clearing
$c$. Overwriting through a residual connection means emitting
$\text{new}-\text{old}$; the subtracted term is linear in the MLP input, so the
result is a ReLU MLP with sufficient depth for the composition of $\phi_k$
and the next message functions. Its content weights do not refer to an
identifier, a node index, or $n$.
\end{proof}

Both endpoint slots of every token are refreshed in the same layer. This is
the step that a single ``aggregate into the target slot'' layer omits: without
refreshing $s_i$, the next layer would aggregate round-$(k-1)$ source states
into round-$(k+1)$ targets.

\subsection{Multiple layers and rounds}

\begin{theorem}\label{thm:main}
For every time-respecting MPNN of Definition~\ref{def:mpnn} with $L$ rounds
there is an edge-only causal transformer with $L$ layers, four identifier-routed
heads per layer, and content width $D=2d_X+d_e+6d_m$, whose content weights
do not depend on the identifiers, node set, or stream length, such that for every
stream $\stream$ and every $\varepsilon>0$ there is a finite temperature
$\alpha$ for which the output token $i$ satisfies
\[
\big\|s_i-h^{(L)}_{u_i}(i)\big\|\le\varepsilon,\qquad
\big\|d_i-h^{(L)}_{v_i}(i)\big\|\le\varepsilon\qquad\text{for all }1\le i\le n.
\]
In the hard-attention limit the equalities are exact. The identifier dimension
is separate from $D$; exact orthonormality requires it to be at least the
number of distinct nodes in the context. The finite routing strength $\alpha$
can depend on the stream and desired accuracy.
\end{theorem}

\begin{proof}
The input MLP maps $[X_{u_i}\|X_{v_i}\|x_i]$ to
$[X_{u_i}\|X_{v_i}\|x_i\|\psi_1(X_{u_i},x_i)\|\psi'_1(X_{v_i},x_i)\|0]$, which
holds round $0$. Lemma~\ref{lem:step} applied $L$ times gives, in the
hard-attention limit, a layer-$L$ output that holds round $L$, so
$s_i=h^{(L)}_{u_i}(i)$ and $d_i=h^{(L)}_{v_i}(i)$.

For finite $\alpha$, fix a finite stream. The ideal messages are bounded,
and the linear projections and ReLU MLPs are Lipschitz. Choose $B$ to bound
the ideal values. Absorb the four-head concatenation, residual maps, and
projection norms into constants $C_k\ge1$. Lemma~\ref{lem:routing} then gives
an error recurrence
\[
\epsilon_k\le C_k(\epsilon_{k-1}+2Bn e^{-\alpha}),\qquad \epsilon_0=0.
\]
Consequently,
\[
\epsilon_L\le 2Bn e^{-\alpha}
\sum_{k=1}^{L}\prod_{\ell=k}^{L}C_\ell
\le 2BnL e^{-\alpha}\prod_{\ell=1}^{L}C_\ell.
\]
For $B>0$, taking
$\alpha\ge\max\{0,\log(2BnL\prod_\ell C_\ell/\varepsilon)\}$
makes the error at most $\varepsilon$. The $B=0$ routing error is zero.
\end{proof}

\begin{corollary}[Bipartite streams]
If no node is both a source and a target, then $\In_i(u_i)=\Out_i(v_i)=\emptyset$
for every $i$, heads T$'$ and S$'$ are unnecessary, and two heads per layer
suffice. Moreover $i\in\In_i(v_i)$ and $i\in\Out_i(u_i)$, so the match sets are
never empty; dropping the BOS token then replaces $\avg$ by the plain mean.
\end{corollary}

\begin{corollary}[Identifiers enter only through coincidences]\label{cor:inv}
With $W_Q=W_K=0$ in the identifier heads, every attention score equals
$\alpha\,\ind[p_\eta(i)=q_\eta(j)]$. Hence the output of the constructed network
(and of any network of the form \eqref{eq:qk} with orthonormal identifiers) is a
function of the contents $(X_{u_i},X_{v_i},x_i)_i$ and of the four coincidence
relations $\ind[u_i=u_j],\ind[v_i=v_j],\ind[u_i=v_j],\ind[v_i=u_j]$ on $j\le i$
alone. It is therefore invariant under $r\mapsto Or$ for orthogonal $O$ and
under any re-assignment of orthonormal identifiers to nodes, and the same weights
apply to every stream over every node set.
\end{corollary}

\begin{remark}[Near-orthogonal identifiers]\label{rem:coherence}
If the identifiers have coherence $\mu=\max_{a\ne b}|\inn{r_a,r_b}|<1$ instead of
$\mu=0$, the off-set scores in Lemma~\ref{lem:routing} are bounded by
$\alpha\mu$ in absolute value and the mass bound becomes $i\,e^{-\alpha(1-\mu)}$;
Theorem~\ref{thm:main} holds with $\alpha$ scaled by $1/(1-\mu)$. Random
unit-norm identifiers in $\R^{d_P}$ achieve $\mu=O(\sqrt{\log|V|/d_P})$ with high
probability, so exact orthonormality is a convenience of exposition, not a
requirement.
\end{remark}

\begin{remark}[What is and is not shown]
Theorem~\ref{thm:main} is a containment statement: the function class of
edge-only causal transformers with orthonormal identifier routing contains
the message-passing recursion in Definition~\ref{def:mpnn}, with one block
per round and shared content weights. It relies on the specified endpoint
pairings, null-attention entry, and sufficient MLP capacity. It does not say
that next-event training recovers these weights, and it makes no claim about
sample efficiency or scaling.
\end{remark}

%% file: editorial/tables/transfer_full.tex
\begin{table}[t]
\centering
\small
\setlength{\tabcolsep}{4pt}
\caption{\textbf{Full link-prediction comparison on TGB} (test MRR):
all nine baseline configurations and our two training settings.
Per-corpus uses one model per dataset (mean $\pm$ std over three seeds); joint uses one model trained on all four (seed $0$).
SimpleDyG is our implementation following the original paper, on the same backbone.
Bold marks the best result in each column.
OOT denotes an evaluation that did not complete within the available budget.}
\label{tab:transfer_full}
\begin{tabular}{lrrrr}
\toprule
Method & \wiki{} & \lastfm{} & \flight{} & \coin{} \\
\midrule
EdgeBank$_{tw}$   & 0.571 & 0.141 & 0.387 & 0.580 \\
EdgeBank$_\infty$ & 0.495 & 0.114 & 0.167 & 0.359 \\
\midrule
Recency (no parameters)    & 0.817 & 0.166 & 0.833 & 0.783 \\
Popularity (no parameters) & 0.040 & 0.013 & 0.603 & 0.607 \\
TGAT             & 0.562 & 0.231 & 0.713 & 0.667 \\
TGN              & 0.611 & 0.220 & 0.791 & 0.637 \\
CAWN             & 0.733 & 0.257 & OOT   & OOT   \\
TCL              & 0.639 & 0.312 & 0.787 & 0.703 \\
GraphMixer       & 0.519 & 0.249 & 0.772 & 0.760 \\
\midrule
DyGFormer        & 0.780 & \textbf{0.395} & 0.843 & 0.747 \\
SimpleDyG        & 0.583 & 0.184 & 0.785 & 0.534 \\
\midrule
Ours (per-corpus) & \textbf{0.823}\,{\scriptsize$\pm$0.000} & 0.333\,{\scriptsize$\pm$0.002} & 0.884\,{\scriptsize$\pm$0.001} & \textbf{0.793}\,{\scriptsize$\pm$0.002} \\
Ours (joint)     & 0.822 & 0.213 & \textbf{0.892} & \textbf{0.793} \\
\bottomrule
\end{tabular}
\end{table}

%% file: editorial/tables/breakdown.tex
\begin{table}[t]
\centering
\small
\setlength{\tabcolsep}{5pt}
\caption{\textbf{Link prediction on recurrent and novel queries} (test MRR,
seed-$0$ per-corpus models, same queries and candidates as
Table~\ref{tab:transfer}). A query is recurrent when its true destination is in
the source's recency list at query time ($256$ slots, $1{,}024$ on \coin{}), and
novel otherwise. Recency is the no-parameter control of Table~\ref{tab:transfer},
which ranks each candidate by the time of its latest event with the source.}
\label{tab:breakdown}
\begin{tabular}{lrrrrr}
\toprule
& & \multicolumn{2}{c}{Recurrent} & \multicolumn{2}{c}{Novel} \\
\cmidrule(lr){3-4}\cmidrule(lr){5-6}
Dataset & Recurrent (\%) & Recency & \ours{} & Recency & \ours{} \\
\midrule
\wiki{}   & 89.2 & 0.916 & \textbf{0.921} & 0.002 & \textbf{0.013} \\
\lastfm{} & 87.6 & 0.189 & \textbf{0.381} & 0.002 & \textbf{0.003} \\
\flight{} & 93.5 & 0.872 & \textbf{0.931} & \textbf{0.278} & 0.183 \\
\coin{}   & 79.4 & 0.913 & 0.913 & 0.283 & \textbf{0.317} \\
\bottomrule
\end{tabular}
\end{table}

%% file: editorial/tables/logo.tex
\begin{table}[t]
\centering
\small
\setlength{\tabcolsep}{4pt}
\caption{\textbf{Transfer to an unseen graph} (test MRR). \emph{Leave-one-out}
trains one model on the other three graphs and evaluates it zero-shot on the
held-out graph, which it never sees during training; \emph{joint} trains on all
four; \emph{per-corpus} trains on the target graph only. Recency is the
no-parameter control of Table~\ref{tab:transfer}. Per-corpus: mean over three seeds; seed $0$ otherwise; joint and
leave-one-out models use $10$k steps and one slot capacity ($1{,}024$).}
\label{tab:logo}
\begin{tabular}{lrrrr}
\toprule
Model & \texttt{tgbl-wiki} & \texttt{tgbl-lastfm} & \texttt{tgbl-flight} & \texttt{tgbl-coin} \\
\midrule
Recency (no parameters)          & 0.817 & 0.166 & 0.833 & 0.783 \\
\ours{} per-corpus               & 0.823 & 0.333 & 0.884 & 0.793 \\
\ours{} joint (all four)         & 0.822 & 0.213 & 0.892 & 0.793 \\
\ours{} leave-one-out (zero-shot) & 0.805 & 0.151 & 0.888 & 0.766 \\
\bottomrule
\end{tabular}
\end{table}

%% file: editorial/tables/graphstats.tex
\begin{table}[t]
\centering
\small
\setlength{\tabcolsep}{4pt}
\caption{\textbf{Dataset statistics.}
Src/Dst count distinct source/destination nodes. Recur. is the fraction of events
whose pair appeared earlier. Times counts distinct timestamps; Days is the span.
Bip. indicates disjoint source/destination node sets. Cand. counts the positive
and evaluation negatives. Each dataset has one edge type.}
\label{tab:graphstats}
{%
\begin{tabular}{lrrrrrrcr}
\toprule
Corpus & Events & Src & Dst & Recur. & Times & Days & Bip. & Cand. \\
\midrule
  \texttt{tgbl-wiki}   & 157K & 8.2K & 1.0K & 0.884 & 153K & 31 & yes & 1000 \\
  \texttt{tgbl-lastfm} & 1.3M & 980 & 1.0K & 0.880 & 1.3M & 1,587 & yes & 1000 \\
  \texttt{tgbl-flight} & 67M & 17K & 18K & 0.966 & 1.4K & 1,399 & no & 21 \\
  \texttt{tgbl-coin}   & 23M & 613K & 638K & 0.831 & 1.3M & 214 & no & 21 \\
\bottomrule
\end{tabular}
}
\end{table}

%% file: editorial/tables/corpora.tex
\begin{table}[t]
\centering
\small
\caption{\textbf{Exact counts for the four datasets.}
Users and items denote source and destination roles. Recurrence measures
previously observed source--destination pairs.}
\label{tab:corpora}
\begin{tabular}{lrrrrr}
\toprule
Corpus & Events & Users & Items & Recurrence & Edge types \\
\midrule
  \texttt{tgbl-wiki}   & 157,474 & 8,227 & 1,000 & 0.8841 & 1 \\
  \texttt{tgbl-lastfm} & 1,293,103 & 980 & 1,000 & 0.8801 & 1 \\
  \texttt{tgbl-flight} & 67,169,570 & 16,706 & 18,143 & 0.9656 & 1 \\
  \texttt{tgbl-coin}   & 22,809,486 & 612,852 & 638,486 & 0.8307 & 1 \\
\bottomrule
\end{tabular}
\end{table}

%% file: editorial/tables/answerable.tex
\begin{table}[t]
\centering
\small
\caption{\textbf{Test-set coverage at recency cap $256$.}
$P(\mathrm{ans})$ is the fraction of destinations in the source's retained list.
Cold destinations have no prior activity; contextual destinations are outside
the list but have prior activity. The three categories partition the split.}
\label{tab:answerable}
\begin{tabular}{lrrr}
\toprule
Corpus & $P(\text{ans})$ & Cold & Contextual \\
\midrule
  \texttt{tgbl-wiki}   & 0.8920 & 0.0007 & 0.1073 \\
  \texttt{tgbl-lastfm} & 0.8761 & 0.0000 & 0.1239 \\
  \texttt{tgbl-flight} & 0.9364 & 0.0000 & 0.0636 \\
  \texttt{tgbl-coin}   & 0.7452 & 0.0080 & 0.2468 \\
\bottomrule
\end{tabular}
\end{table}

%% file: editorial/tables/params.tex
  \begin{table}[t]
  \centering
\small
  \caption{\textbf{Trainable parameter counts.}
The vocabulary-free row describes the supplied fixed reference configuration.
Other run configurations, including the joint model, can differ.
Non-trainable histories and memory buffers are excluded.}
  \label{tab:params}
  \begin{tabular}{lrrr}
  \toprule
  Model & \texttt{tgbl-wiki} & \texttt{tgbl-lastfm} & \texttt{tgbl-flight} \\
  \midrule
    DyRep       & $0.69$M & $0.57$M & $0.58$M \\
    TGAT        & $0.65$M & $0.52$M & $0.53$M \\
    TGN         & $0.96$M & $0.78$M & $0.80$M \\
    CAWN        & $2.24$M & $1.06$M & $1.11$M \\
    TCL         & $0.86$M & $0.83$M & $0.83$M \\
    GraphMixer  & $0.61$M & $0.58$M & $0.59$M \\
    DyGFormer   & $1.08$M & $1.07$M & $1.07$M \\
    SimpleDyG & $0.94$M & $0.94$M & $3.15$M \\
  \midrule
    \textbf{Ours (vocab-free)} & $\mathbf{0.93}$\textbf{M} & $\mathbf{0.93}$\textbf{M} & $\mathbf{0.93}$\textbf{M} \\
  \bottomrule
  \end{tabular}
  \end{table}

%% file: editorial/tables/runtime.tex
\begin{table}[t]
\centering
\small
\caption{\textbf{Recorded training-time estimates on an L40S GPU.}
Evaluation is excluded. Values are retained from
the source. OOM indicates an allocation failure; OOT indicates an
unfinished evaluation. Bold identifies our row, not the fastest entry.}
\label{tab:runtime}
{%
\begin{tabular}{lrrrr}
\toprule
Model & \texttt{tgbl-wiki} & \texttt{tgbl-lastfm} & \texttt{tgbl-flight} & \texttt{tgbl-coin} \\
\midrule
  TGN         & $6.2$ min & $40.8$ min & $13.7$ h & $2.24$ h \\
  GraphMixer  & $12.1$ min & $54.1$ min & $6.89$ h & $3.65$ h \\
  TCL         & $13.6$ min & $2.21$ h & $7.20$ h & $2.73$ h \\
  TGAT        & $31.4$ min & $2.57$ h & $25.5$ h & $4.99$ h \\
  CAWN        & $1.09$ h & $4.08$ h & OOT & OOT \\
  DyGFormer   & $15.2$ min & $1.54$ h & $5.01$ h & $2.61$ h \\
\midrule
  \textbf{\ours{} (per dataset)} & $\mathbf{13.9}$\textbf{ min} & $\mathbf{13.5}$\textbf{ min} & $\mathbf{34.6}$\textbf{ min} & $\mathbf{24.3}$\textbf{ min} \\
    \textbf{\ours{} (joint)}
    & \multicolumn{4}{c}{$\mathbf{47.7}$\textbf{ min for one model trained on all four datasets}} \\
\bottomrule
\end{tabular}
}
\end{table}

%% file: editorial/tables/cost.tex
\begin{table}[t]
\centering
\small
\caption{\textbf{Inference throughput on an idle NVIDIA L40S.}
Positives ranked per second, including the stated evaluation negatives.
Baseline batch sizes are $32$ on \wiki{}/\lastfm{} ($16$ for CAWN) and $200$
on the larger datasets. Slot capacity also differs: the \coin{} run uses $1{,}024$.
Compare methods within a dataset; these are implementation-level measurements.}
\label{tab:cost}
{%
\begin{tabular}{lrrrr}
\toprule
Method & \texttt{tgbl-wiki} & \texttt{tgbl-lastfm} & \texttt{tgbl-flight} & \texttt{tgbl-coin} \\
\midrule
Negatives per positive & $999$ & $999$ & $20$ & $20$ \\
\midrule
  DyRep       & $110$ & $109$ & $3{,}732$ & $669$ \\
  TGAT        & $4.8$ & $5.2$ & $230$ & $254$ \\
  TGN         & $113$ & $114$ & $3{,}787$ & $670$ \\
  CAWN        & $2.6$ & $2.1$ & $11.9$ & $26.8$ \\
  TCL         & $44$ & $45$ & $2{,}052$ & $2{,}058$ \\
  GraphMixer  & $9.9$ & $12.8$ & $638$ & $533$ \\
  DyGFormer   & $22$ & $30$ & $778$ & $832$ \\
  \textbf{Ours (vocab-free)} & $\mathbf{930}$ & $\mathbf{896}$ & $2{,}295$ & 898 \\
\bottomrule
\end{tabular}
}
\end{table}

%% file: editorial/tables/ablation.tex
\begin{table}[t]
\centering
\small
\setlength{\tabcolsep}{6pt}
\caption{\textbf{Component ablations} (test MRR, seed $0$, $6$k training steps).
Each variant removes one component from the per-corpus model. \emph{Local
identifiers}: the identifier attention bias is set to zero. \emph{Descriptor
prediction}: the descriptor head outputs a uniform distribution, so novel
candidates are ranked only by $p_{\mathrm{nov}}$ and $|\mathcal{C}_t(v)|$.
\emph{Recency pointer}: the pointer output is removed and every candidate is
scored by the descriptor term of Equation~\ref{eq:main_score}. \emph{Slot-bias
head}: the per-slot bias from history statistics is removed. \lastfm{} uses a $50$k-query test subset.}
\label{tab:ablation}
\begin{tabular}{lrr}
\toprule
Model & \wiki{} & \lastfm{} \\
\midrule
\ours{} (full)                & 0.823 & 0.251 \\
\quad $-$ local identifiers   & 0.823 & 0.247 \\
\quad $-$ descriptor prediction & 0.823 & 0.244 \\
\quad $-$ recency pointer     & 0.472 & 0.092 \\
\quad $-$ slot-bias head      & 0.819 & 0.229 \\
\bottomrule
\end{tabular}
\end{table}

%% file: editorial/tables/negsweep.tex
\begin{table}[t]
\centering
\small
\caption{\textbf{Candidate-width sweep.} Test MRR of the seed-$0$ per-corpus models of
Table~\ref{tab:transfer} with $20$, $100$, or all $999$ official negatives per
positive. Smaller widths keep a random subset of the official negatives.}
\label{tab:negsweep}
\begin{tabular}{lrr}
\toprule
Candidates & \texttt{tgbl-wiki} & \texttt{tgbl-lastfm} \\
\midrule
  $21$      & 0.916 & 0.724 \\
  $101$     & 0.884 & 0.543 \\
  $1{,}000$ & 0.823 & 0.334 \\
\bottomrule
\end{tabular}
\end{table}

%% file: editorial/tables/event_gap_full.tex
\begin{table}[t]
\centering
\small
\setlength{\tabcolsep}{4pt}
\caption{\textbf{Next-event time-gap prediction} (NLL in nats, lower is better /
bucket accuracy \%, $13$ buckets; $10{,}000$ test queries per dataset, seed $0$).
Baselines and \ours{} use the same source history up to $t_{\text{last}}$ and
have no access to the query time; baseline weights are chosen on validation queries. Random reports its expected accuracy ($1/13$).
\ours{} variants: without the gap-history head (\wiki{} $6$k, \lastfm{} $20$k,
\coin{} $10$k steps) and with it (\wiki{} and \lastfm{} $6$k, \coin{} $10$k steps).
\flight{} is omitted: its timestamps have day resolution, so $99.7\%$ of targets
fall in one bucket (1 day--1 week).}
\label{tab:event_gap_full}
\begin{tabular}{lrrrrrr}
\toprule
& \multicolumn{2}{c}{\wiki{}} & \multicolumn{2}{c}{\lastfm{}} & \multicolumn{2}{c}{\coin{}} \\
\cmidrule(lr){2-3}\cmidrule(lr){4-5}\cmidrule(lr){6-7}
Method & NLL & Acc. & NLL & Acc. & NLL & Acc. \\
\midrule
Random                          & 2.565 &  7.7 & 2.565 &  7.7 & 2.565 &  7.7 \\
Majority                        & 1.920 & 28.4 & 2.264 & 28.9 & 1.836 & 29.5 \\
Persistence (last gap)          & 1.878 & 27.3 & 1.742 & 31.4 & 1.719 & 39.9 \\
Poisson process (per user)      & 1.896 & 28.4 & 1.920 & 28.9 & 1.620 & 38.7 \\
Per-user empirical distribution & 1.760 & 33.3 & 1.482 & 43.6 & 1.352 & 47.9 \\
\midrule
\ours{}                         & 1.724 & 33.2 & 1.496 & 39.8 & 1.369 & 47.9 \\
\ours{} + gap-history head      & \textbf{1.706} & \textbf{33.7} & \textbf{1.420} & \textbf{46.5} & \textbf{1.320} & \textbf{49.6} \\
\bottomrule
\end{tabular}
\end{table}